\documentclass{article}

     \PassOptionsToPackage{numbers, compress}{natbib}

 \usepackage[preprint]{neurips_2021}

\usepackage[utf8]{inputenc} 
\usepackage[T1]{fontenc}    
\usepackage{hyperref}       
\usepackage{url}            
\usepackage{booktabs}       
\usepackage{amsfonts}       
\usepackage{nicefrac}       
\usepackage{microtype}      
\usepackage{fancyhdr,graphicx,amsmath,amssymb}
\usepackage{algorithm}
\usepackage{algorithmic}
\usepackage{amsmath}
\usepackage{xcolor}
\usepackage{comment}
\title{Temporal Early Exits for Efficient Video Object Detection}

\author{%
  Amin Sabet\\
  School of Electronics and Computer Science\\
  University of Southampton, UK\\
  \texttt{ms4r18@soton.ac.uk} \\
   \And
   Jonathon Hare \\
   School of Electronics and Computer Science\\
  University of Southampton, UK\\
  \texttt{jsh2@ecs.soton.ac.uk,} \\
   \AND
   Bashir Al-Hashimi \\
   Faculty of Natural and Mathematical Sciences\\
   King’s College London, UK \\
   \texttt{bashir.al-hashimi@kcl.ac.uk} \\
   \And
   Geoff V. Merrett \\
   School of Electronics and Computer Science\\
  University of Southampton, UK\\
  \texttt{gvm@ecs.soton.ac.uk} \\
}

\begin{document}

\maketitle

\begin{abstract}
Transferring image-based object detectors to the domain of video remains challenging under resource constraints. Previous efforts utilised optical flow to allow unchanged features to be propagated, however, the overhead is considerable when working with very slowly changing scenes from applications such as surveillance. In this paper,  we propose temporal early exits to reduce the computational complexity of per-frame video object detection. Multiple temporal early exit modules with low computational overhead are inserted at early layers of the backbone network to identify the semantic differences between consecutive frames. Full computation is only required if the frame is identified as having a semantic change to previous frames; otherwise, detection results from previous frames are reused.  Experiments on CDnet show that our method significantly reduces the computational complexity and execution of per-frame video object detection up to $34 \times$ compared to existing methods with an acceptable reduction of 2.2\% in mAP.
\end{abstract}

\section{Introduction}
\begin{figure}
  \centering
  \includegraphics[width=0.9\textwidth]{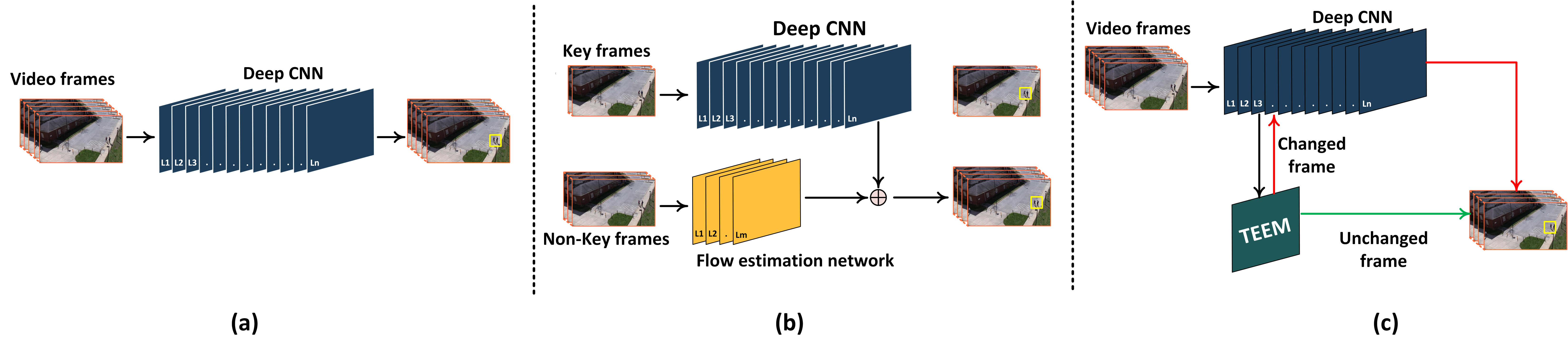}
  \caption{Comparison video object detection approaches. (a) Conventional approach: applying deep CNNs on individual frames. (b) Deep flow estimation: employing lighter flow estimation network to propagate features across frames. (c) Proposed pipeline: identifying semantic variation in early stages of network and avoiding deep CNN computation for unchanged video frames }
  \label{fig1:frameworks}
\end{figure}

Object detection is one of the fundamental tasks in computer vision and serves as a core approach in many practical applications, such as robotics and video surveillance \cite{tian2015deep,borji2015salient}. Object detection in static images has achieved remarkable successes in recent years using CNNs \cite{huang2017speed}. However, video object detection has now emerged as a new challenge beyond image data. This is due to the high computational cost introduced by applying existing image object detection networks on numerous individual video frames. Figure \ref{fig1:frameworks}-(a) shows the overview of the per-frame video object detection approach, where all video frames are processed by a similar CNN. Deploying per-frame video object detection becomes even more challenging for resource and energy-constrained applications.  

Deep optical flow approaches \cite{deepflow} tackle the computational complexity challenge of video object detection by taking advantage of temporal information in videos. They exploit feature similarity between consecutive frames to reduce the expensive feature computation on most frames and improve the speed. Instead of extracting features of all frames by a deep CNN, deep optical flow uses a lighter network to extract, propagate and aggregate features of video frames with similar features to previous frames. Figure \ref{fig1:frameworks}-(b) shows the overview of deep optical flow approaches \cite{shidifferential}.  

Feature similarity between successive video frames occurs often in applications such as facility monitoring and surveillance systems, where the camera is static and there are less frequently moving objects in the videos \cite{luo2018key}. Whilst reducing computational complexity, optical flow approaches require substantial computational effort to generate and aggregating feature maps even though the features of successive frames remain unchanged.

To address the challenge of identifying and processing frames with unchanged features, we propose a computationally lightweight, Temporal Early Exit Module (TEEM), which identifies semantic variations between consecutive frames. We show that the full computational effort of a network \cite{park2019robust} is not required to distinguish and detect semantic changes between frames. We then use TEEM to build a per-frame video object detection pipeline, shown in figure \ref{fig1:frameworks}-(C). In this pipeline, a TEEM identifies semantic variation between features of consecutive frames in the very early stages of the feature network. Then, the TEEM conditionally activates the deeper layers of the network if a semantic difference is detected between frames. If a frame is identified to be semantically unchanged, then the object detection results from the previous frame are reused. By identifying unchanged frames at earlier stages and thus, avoiding processing video frames, the proposed pipeline significantly reduces the computational complexity and speeds up the video object detection. 

The contributions of this paper are as follows.
\begin{itemize}
  \item \textbf{We propose Temporal Early Exit Modules which exploit the features of the early convolutional layers to infer a 2D spatial attention map between video frames.} The attention map encodes the semantic variations between consecutive frames. The attention map is then used to generate refined feature maps, feeding into a classifier to classify frames into changed and unchanged categories. The experiments on CDnet dataset \cite{CDnet} show that the TEEM classifies frames into changed and unchanged classes with 94\% accuracy. 
  
  \item \textbf{We demonstrate a temporal early exit video object detection pipeline that uses TEEMs to conditionally process video frames.} Full computation effort is required only for processing video frames with a semantic variation to previous frames. The evaluation of the proposed pipline on CDnet dataset \cite{CDnet} shows up to $34 \times$ speed up of per-frame video object detection with less than 2.2\% reduction in mAP.  
\end{itemize}

\section{Related Work}


\paragraph{Optical Flow Estimation.}
Recent optical flow estimation approaches divide all video frames into key frame and non-key frame sets \cite{zhu2017deep} \cite{shidifferential} \cite{deepflow}. While deep networks are only applied on the key-frames, a lighter flow estimation network is exploited to obtain the features of non-key frames (shown in Figure \ref{fig1:frameworks}), which results in speeding up the algorithm. However, the flow estimation approaches rely on selecting the best key-frame, where the features of non-key frames are estimated from key-frame features. Deep feature flow network (DFF) \cite{zhu2017deep} adopts a simple fixed key-frame selection scheme and fails to take account of the quality of key-frame and the optical-flow. The differential network \cite{shidifferential} proposes a binary differential classifier network to detect the key-frames with classification accuracy of up to 76\%. In order to address the challenge of finding the optimal key video frames, instead of replacing the temporal features with the features of key-frames, PSLA \cite{guo2019progressive} proposes to progressively update the temporal features through a recursive feature updating network. To summarise, while flow estimation networks speed up video object detection, they require large and redundant computations for feature propagation for the majority of unchanged video frames.

 \begin{figure*}
     \centering
     \includegraphics[width=13cm]{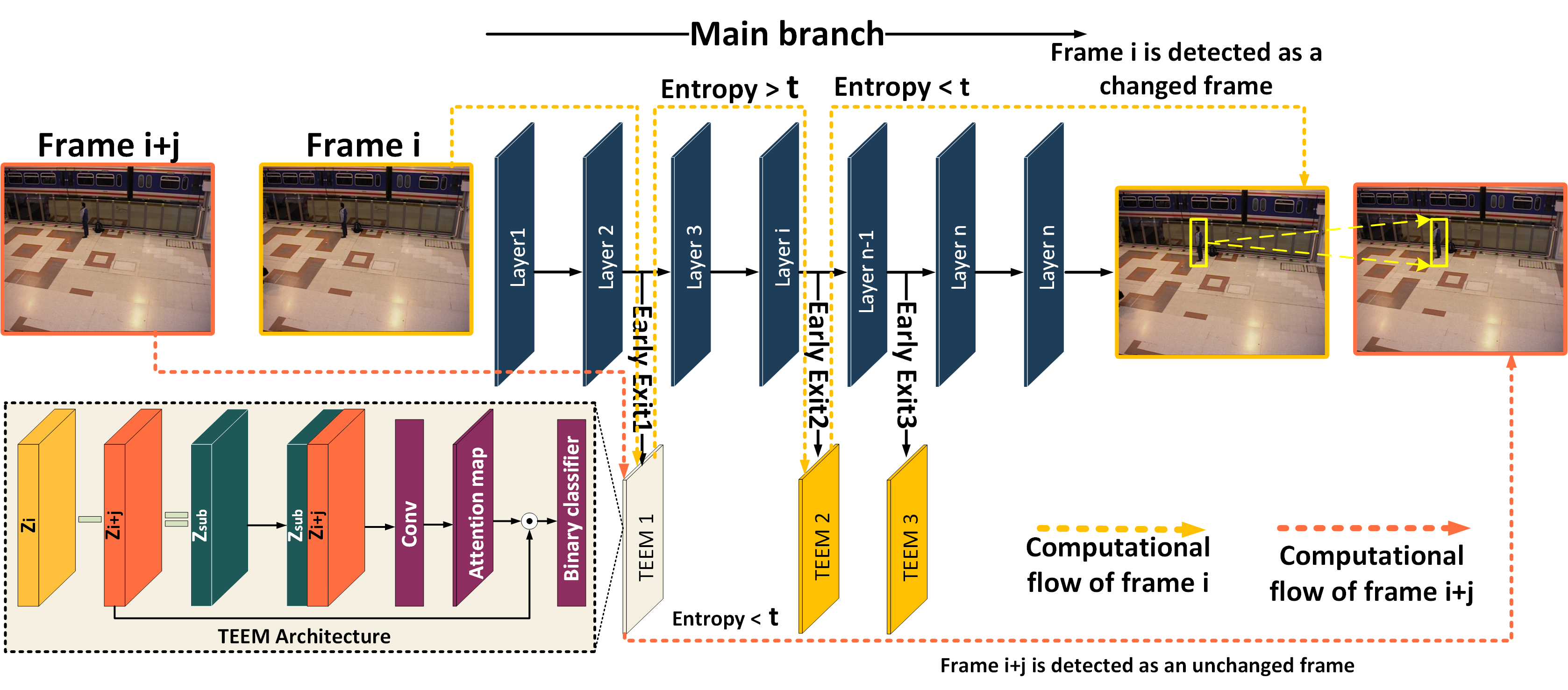}
     \caption{Overview of the proposed temporal early exits video object detection pipeline.TEEMs are added to feature network of a object detection network (main branch) to detect unchanged video frames and avoid redundant computations.}
     \label{fig:overview}
 \end{figure*}

\paragraph{Early Exit CNNs.} In a similar line of work to ours, early exit classifiers promote faster inference by allowing classification for easy instances to exit the network early. This class of algorithms are based on the observation that  often the features learned at earlier stages of a deep network can correctly infer a large subset of the data samples. For example, BranchyNet \cite{branchynet} is a neural network architecture which adds side branches to the main branch, the original backbone network, to allow certain test samples to exit early. BranchyNet is trained by the joint optimization of loss functions for all exit points. 
In the follow-up work, conditional deep learning (CDL) \cite{conditional} measures the efficiency improvement due to the addition of the linear classifier at each convolutional layer to find the optimal layer for early exit branches in the backbone network. Early exit approaches reduce the runtime and energy consumption of image classification. However, they have been developed only for image classification applications.


\textbf{Scene Change Detection.} Scene change detection is a fundamental problem in the field of computer vision. The most  classical approach for scene change detection is image differencing~\cite{scd17,scd18}, which generates a change map by determining the set of pixels that are `significantly different' across two images. Then, a binary mask is generated by thresholding the change map. Although, this method has low computational cost, the raw pixel features are incapable of effectively differentiating between semantic changes and noise. To obtain more discriminative features, image rationing~\cite{scd19}, change vector analysis~\cite{scd20,scd21}, Markov random fields~\cite{scd22} and dictionary learning~\cite{scd19,scd23} have been proposed. However, they are still sensitive to noise and illumination changes~\cite{svd16}.

 \begin{figure*}[]
 \footnotesize
\begin{tabular}{cccccc}
    (A) &  \includegraphics[width=.14\linewidth]{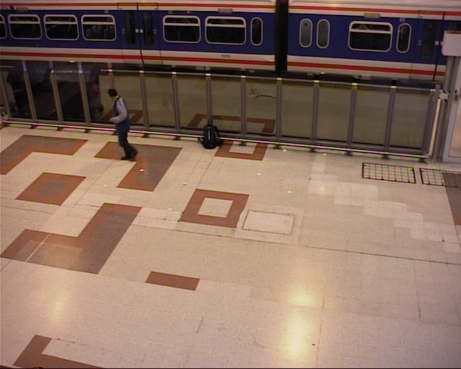} &       \includegraphics[width=.14\linewidth]{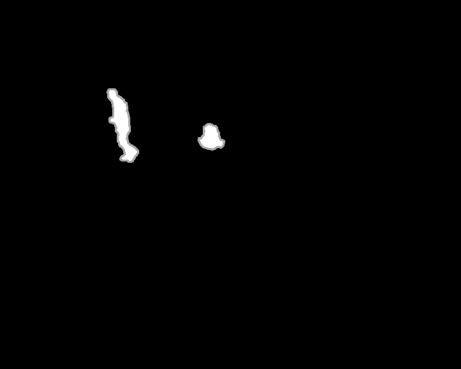} &       \includegraphics[width=.14\linewidth]{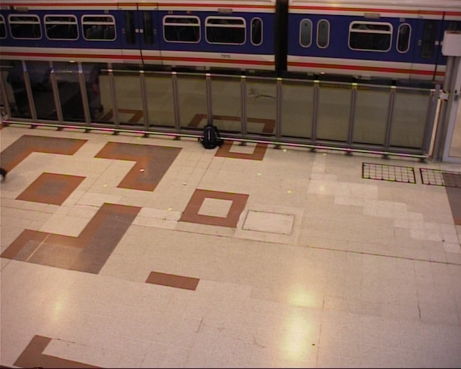} &       \includegraphics[width=.14\linewidth]{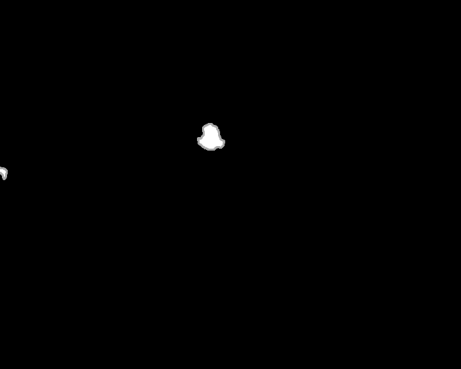} &     \includegraphics[width=.14\linewidth]{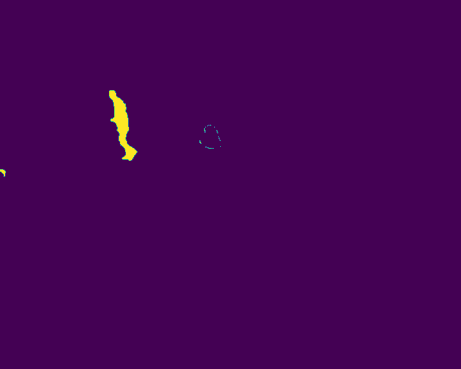}    \\
(B) &    \includegraphics[width=.14\linewidth]{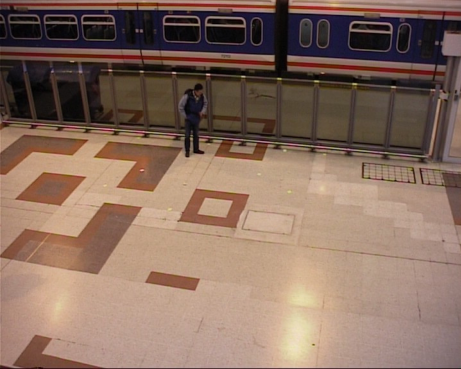} &       \includegraphics[width=.14\linewidth]{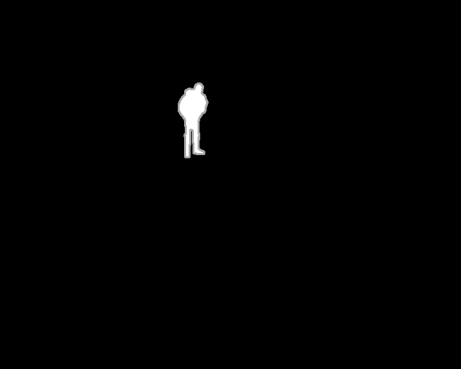} &       \includegraphics[width=.14\linewidth]{8.png} &       \includegraphics[width=.14\linewidth]{Seg_8.png} &     \includegraphics[width=.14\linewidth]{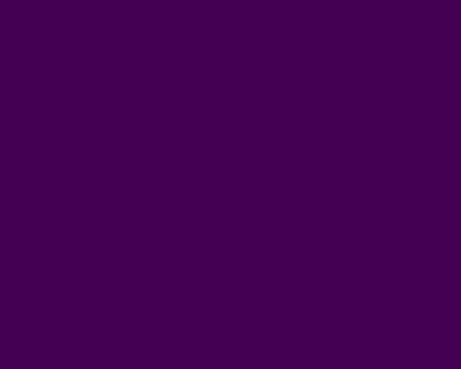}    \\
  (C) &  \includegraphics[width=.14\linewidth]{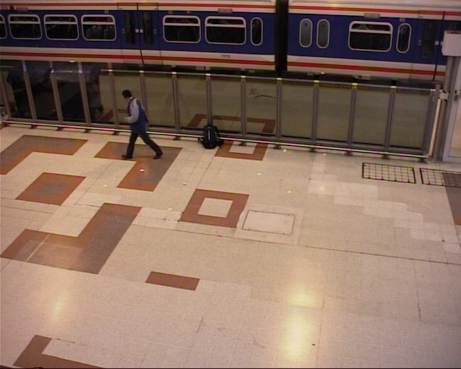} &       \includegraphics[width=.14\linewidth]{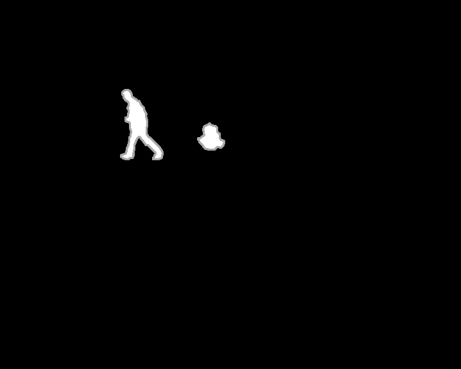} &       \includegraphics[width=.14\linewidth]{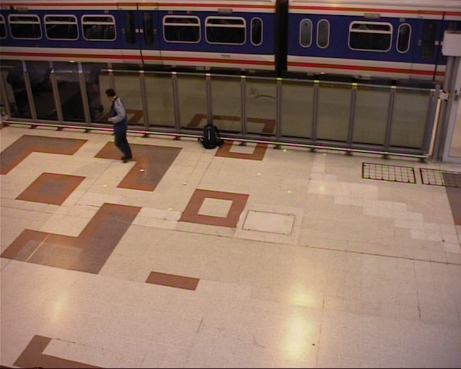} &       \includegraphics[width=.14\linewidth]{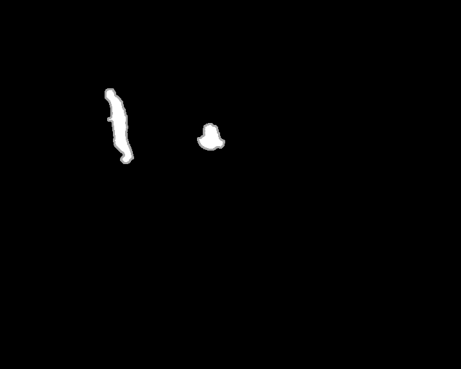} &     \includegraphics[width=.14\linewidth]{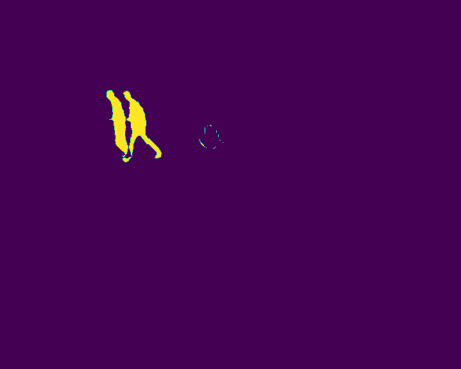}    \\
   &    $fr_{i}$ & objects of interest i & $fr_{i+j}$ & objects of interest i+j & $MFI(fr_i,fr_{i+j})$
\end{tabular}

\caption{Scenery change examples, (A) shows a changed scenery with the IoU=0. (B) shows an unchanged scenery with the IoU=1 (C) shows a changed scenery with the IoU=0.19}
\label{fig:sc}
\end{figure*}

\section{Temporal Early Exits Video Object Detection Pipeline}

We propose a novel video object detection pipeline to tackle the computational complexity of video object detection. An overview of the proposed pipeline is shown in figure \ref{fig:overview}. Given a sequence of video frames, multiple Temporal Early Exit Modules (TEEM) are used to build the early exit branches. Each early exit branch uses features extracted in the early stages of the feature network to identify any semantic variation between the features of consecutive video frames. If any semantic variations is distinguished by a TEEM, a full computation effort (main branch) will be needed to process the video frame. On the other hand, the computation process of a video frame without any semantic variations will be terminated at an early exit branch and the detection results from the previous frame will be reused. We describe the implementation details of the proposed pipeline in this section. 

\subsection{Scenery Change}
 The key challenge in identifying feature variations is qualifying semantic dissimilarities between images \cite{alcantarilla2018street}. To address this challenge, we introduce a scenery change $\text(SC)$ metric to quantify semantic variations between video frames. We define a semantic variation to be only a variation in objects of interest where noisy variations are considered to be nuisance. Given any pairs of video frames, a TEEM aims to identify semantic differences between frames. A scenery changes between a pair of video frames $(fr_i,fr_{i+j})$, if the maximum motion in the Moving Fields of Interest (MFI) is greater than the variation threshold value $\tau_{var}$ as follows,
 \begin{equation}
 \footnotesize \text{SC}(fr_i,fr_{i+j})= 
\begin{cases}
    1&  \max (\text{MFI}(fr_i,fr_{i+j}))>  \tau_{var} \\
    0              & \text{otherwise}
\end{cases}.
\label{eq:SC}
 \end{equation}
 
  We compute MFI set by measuring intersection over union (IoU) of each object of interest, $O_k$, across frame pairs as follows,
 
  \begin{equation}
 \footnotesize
         \text{MFI}(o_k)= 
\begin{cases}
    1-\text{IoU}(O_{k}^{fr_i},O_{k}^{fr_{i+j}})& \text{if } O_{k} \in fr_i \land fr_{i+j} \\
    1              & \text{Otherwise}
\end{cases}.
 \end{equation}
 Each element in MFI quantifies the variation in an object of interest between a pair of video frames. The scenery change metric distinguishes noisy variations from semantic variations in video frame pairs. Figure \ref{fig:sc} visualizes some examples of scenery change metric for video frame pair.
 
\subsection{Temporal Early Exit Module}
\label{sec:TEEM}
To detect a scenery change between frame pairs $(fr_i,fr_{i+j})$, we propose TEEM which consists of \textit{Attention} and \textit{Classifier} components. This section describes these components in details. Figure \ref{fig:overview} shows an overview of a TEEM.

The $Attention$ component is represented as a function parameterized by $ {\theta _{\text{Atten}}}_{l}$ in equation \eqref{eq:att}.  $F_{\text{Attn}}$ takes $Z_{i}^{l}$ and $Z_{i+j}^{l}$ as inputs, and  outputs an attention map $ \text{AttMap}_l$ which encodes the semantic variations happened between $fr_{i}$ and $fr_{i+j}$. In equation \eqref{eq:att}, $Z_{i}^{l}$ and $Z_{i+j}^{l}$ are image features of the previous $fr_i$ and current $fr_{i+j}$ video frames, respectively, encoded by convolutional layer $l$ in the feature network. The attention function captures the semantic variations between frame pairs in the representation space by subtracting $Z_{i}^{l}$ from $Z_{i+j}^{l}$ in equation \eqref{eq:sub}. Then, by concatenating the subtraction results with the feature representation of the current video frame, $Z_{i+j}^{l}$, a 2D spatial attention map associated with the video frame pair is generated in equation \eqref{eq:sig}.

 \begin{equation}
 \label{eq:att}
 \footnotesize
     {\text{AttMap}}_{l} = F_{\text{Attn}}(Z_{i}^{l},Z_{i+j}^{l}; {\theta _\text{Atten}}_{l}).
 \end{equation}
\begin{equation}
 \footnotesize
    Z_{\text{sub}} = Z_{i+j}-Z_{j},
    \label{eq:sub}
\end{equation}
\begin{equation}
 \footnotesize
    \bar{Z}_{c}=\text{Conct}[Z_i:Z_{sub}],
    \label{eq:conct}
\end{equation}
\begin{equation}
\footnotesize
    \text{AttMap}_{l} = \text{Sigmoid} (\text{Conv}( \bar{Z}_{c})),
    \label{eq:sig}
\end{equation}
In equations \eqref{eq:sig} and \eqref{eq:conct}, $\text{Conct}[;]$, $\text{Conv}$, $\text{Sigmoid}$ indicate concatenation, convolutional layer and \textit{sigmoid} function, respectively. Following \cite{mascharka2018transparency}, to avoid introducing any form of global normalization, we use sigmoid, instead of softmax, for computing the attention map. The attention map makes TEEM learn and focus on the motions of objects of interest between frame pairs. Notably, a TEEM requires storing the latest feature maps $Z_{i}^{l}$ to be used for successive frames. 

The generated attention map alongside the feature map of the current video frame $Z_{i+j}^{l}$ are fed to a $classifier$ to detect any scenery change between frame pairs. The $classifier$ component is a classifier with two outputs which represented as a function $F_{\text{Class}}$, parameterized by $ \theta _{\text{Class}}$ in equation \eqref{eq:classifier}. It takes the attention map, $\text{AttMap}_l$, and the feature map of the current video frame $Z_{i+j}^{l}$ and produces a class label. The possible labels are changed and unchanged frame. To classify the video frame, the $\text{AttMap}_l$ is multiplied element-wise with the feature map of current video frame to get the refined feature maps $ Z_{\text{Att}}$ in equation \eqref{eq:attion}. The refined feature maps are passed through a convolutional layer and a fully connected layer, equation \eqref{eq:FC_layer}, to produce a class label.

 \begin{equation}
 \footnotesize
     \text{Cls}^{l}_{i}  = F_{\text{Class}}(Z_{i+j}^{l}, \text{AttMap}_l; \theta _{\text{Class}}).
     \label{eq:classifier}
 \end{equation}
\begin{equation}
\footnotesize
    Z_{\text{Att}} =  \text{AttMap}_l \odot Z_{i+j}^{l}.
    \label{eq:attion}
\end{equation}
\begin{equation}
\footnotesize
    \text{Cls}^{l}_{i} =  \text{FC}(\text{ReLU}(\text{Norm}(\text{Conv}(Z_{\text{att}})))).
    \label{eq:FC_layer}
\end{equation}
 In equation \eqref{eq:FC_layer} FC, ReLU, Norm and Conv indicate fully connected, ReLU, batch normalization and convolutional layers, respectively.  $\text{Cls}^{l}_{i}$ is the output of the TEEM, located at the convolutional layer $l$ of feature network. Each video frame can be classified by a TEEM located in an early exit branch into a changed or unchanged pair.

\subsection{Training Policy}
 An early exit branch, including a TEEM, can be added after any convolutional layer in the feature network. However, the location of an early exit branch in the feature network determines its performance in identifying semantic variations. Therefore, a temporal early attention video object detection pipeline have multiple early exits to identify semantic variations. 
 However, each TEEM in an early exit branch is regarded as a stand-alone module, hence, trained separately from the backbone network. Let $(fr_{i},fr_{i+j})$ be a sampled video frame pair and $ \theta$ be all parameters in $\text{TEEM}$ in early exit $l$. For a target label, $P$, the objective is to minimize the cross-entropy loss as follows:

\begin{equation}
    {L_\text{TEEM}}_{k}(P,P_{\theta};\theta)=-\sum_{k}P_{k} \cdot  log({P_{\theta}}_{k}) 
\end{equation}
where $k$ is classifier outputs and,
\begin{equation}
 \footnotesize
    P_{\theta} = \text{Softmax}(S),
\end{equation}
and
\begin{equation}
 \footnotesize
    S = F_{\text{TEEM}_{l}}(fr_{i},fr_{i+j};\theta).
\end{equation}
Here, $F_{\text{TEEM}_{l}}$ is the the output of the TEEM in early exit $l$. In the forward pass of training, a video frame pair is passed through the network, including both the main branch and early exits. The output of each early exit is recorded to calculate the error of TEEM in the early exit. In backward propagation, the error of each exit is passed back only through the TEEMs and the weights of TEEM are updated using gradient descent. Notably, the weights of the main branch are not updated during training of the TEEMs.

\subsection{Fast inference in a Temporal Early Exit Video Object Detection Pipeline}
Algorithm \ref{alg:inference} summarizes the temporal early exit video object detection pipeline inference algorithm. Given a pre-trained object detection network as the main branch, $F_{\text{main}}$, with $j$ early exit branches, the first video frame is processed by the main branch. The inference for successive frames proceeds by feeding frames through the feature network of the main branch until reaching the first early exit branch. Then, the softmax and entropy of $F_{\text{TEEM}}$ in the early exit are calculated to evaluate the confidence of the classifier in prediction. If the entropy is less than the threshold $\gamma$ and the class label with the maximum score (probability) is unchanged, the inference procedure terminates and object detection results from the previous frame are returned. If the class label with maximum score is changed, then the frame is processed by the main branch. For the entropy greater than $\gamma$, the frame continues to be processed by the feature network till the next early exit branch. If the softmax and entropy of all early exits are greater than thresholds, the frame continues to be processed by the main branch.

\begin{algorithm}
\caption{Inference algorithm for video object detection}
\begin{algorithmic} 
\REQUIRE $\text{Video frames} \left \{ I \right \}  $
\ENSURE $\text{detection} = \text{None}$
\FOR{$i\leftarrow 1$ \textbf{to} $\leftarrow \text{length}(I)$} 
\IF{$i==1$}  
\STATE $ \text{detection} \leftarrow F_{\text{main}}(I_{i})$
\RETURN $\text{detection}$
\ELSE
\FOR{$j\leftarrow1$ \textbf{to} $j=\#(\text{early exits})$}
\STATE $y\leftarrow F_{{\text{TEEM}}_{j}}(I_i,I_{i-1})$
\STATE $S\leftarrow \text{softmax}(y)$
\STATE $e \leftarrow \text{entropy}(S)$
\IF{$e <\tau$ and $\text{arg max}(S) == \text{Unchanged}$}
\RETURN $\text{detection}$
\ELSIF{$e <\gamma$ and $\text{arg max}(S) == \text{Changed}$}
\STATE $\text{detection}  \leftarrow F_{\text{main}}(I_{i})$
\RETURN$\text{detection}$
\ELSIF{$e>\gamma$}
\STATE \textbf{pass}
\ENDIF
\ENDFOR
\ENDIF
\ENDFOR
\end{algorithmic}
\label{alg:inference}
\end{algorithm}

\section{Dataset}

The CDnet  dataset \cite{CDnet} provides a realistic, camera-captured, diverse set of videos for change detection. The videos have been selected to cover a wide range of detection challenges and are representative of typical indoor and outdoor visual data captured in surveillance, smart environment. This dataset consists of 33 camera-captured videos ($\sim$70,000 frames) with boats, trucks, and pedestrians as objects of interest. It contains a range of challenging factors, including turbulence, dynamic background, camera jitter challenging weather and pan-tilt-zooming (PTZ). All videos come with ground-truth segmentation of motion areas for each video frame.
However, CDnet dataset doesn't provide labels for the individual instances of moving objects. This makes measuring SC metric challenging and inaccurate in scenarios with multiple moving objects. To evaluate those scenarios, we also build the scenery change dataset. Our dataset is based on VIRAT dataset \cite{VIRAT} which is inspired and used by many change detection applications \cite{park2019robust}. We used Vatic engine \cite{Vatic} to label individual moving objects with bounding boxes. The dataset consists of 20 camera-captured videos with pedestrian, car and cyclist as objects of interest. We believe our dataset can complement existing change detection datasets like CDnet\cite{CDnet}. 

\subsection{Dataset Balancing}

To train TEEMs, video frame pairs need to be randomly sampled from videos. However, random sampling of frame pairs causes the unbalanced class dataset (changed scenery and unchanged scenery classes). The balance of classes depends on the interval between sampled video frame pairs. For long sampling interval, the dataset skews to changed scenery frame pairs. On the other hand, a short sampling interval leads to skewing to unchanged scenery. As a result of the unbalanced dataset, the classifier of TEEMs tends to ignore small classes, while concentrating on learning to classify large ones accurately. We tackle the unbalanced dataset problem by proposing the dynamic sampling interval approach. To this end, we divide the semantic variation into ten classes. Each class represents a variation threshold between frame pairs i.e., class one and two represent frame pairs with variation threshold of [0, 0.1) and [0.1, 0.2), respectively. Consequently, starting from a random sampling interval, we sample frame pairs, measure the minimum $IoU$ of objects across frame pairs and classify them into one of the ten classes. As the number of pairs in one class exceeds a threshold, we reject the sampling that falls into the class and adjusts the sampling interval toward the direction which increases classes with a smaller number of samples until all classes are balanced. We also face another class unbalance issue due to videos with different lengths --- i.e., some videos have thousands of frames while some have a few dozens of frames. We address this problem by upsampling frame pairs from shorter videos.
 
\section{Experiments and Results}\label{Sec:exper}

In this section, we describe the experimental setup used to evaluate the performance of the temporal early exit video object detection pipeline. The experimental results are also presented in this section.
\subsection{Implementation and Experimental Setup}
 For the main branch of the video object detection pipeline , $F_{\text{main}}$, we use Faster-RCNN \cite{ren2015faster} with Resnet50 and Resnet101 feature networks trained on MS COCO detection dataset \cite{lin2014microsoft}. We use TEE-Faster-RCNN to refer to this pipeline. Following Residual attention network \cite{wang2017residual}, we add early exit branches after each Bottleneck and Basic blocks of the feature network. Therefore, TEE-Faster-RCNN consists of four early exits. We train TEEMs for 40 epochs using the Adam Optimizer \cite{kingma2014adam} with a learning rate of 0.001 and a batch size of 64. In both training and inference, the images have shorter sides of 224 pixels. We use 25K and 5k video frame pairs for training and testing, respectively. Training was performed on 4 GPUs and testing run time is measured on a single GTX 1080 GPU. We set the entropy to 0.97 at test time. Our model is implemented using PyTorch \cite{paszke2019pytorch}, and our code and dataset will be made publicly available. We evaluate the accuracy, computational complexity, and run time of TEE-Faster-RCNN.
 
 \subsection{Classification Accuracy of Early Exits:} 
 We evaluate the accuracy of early exits using classification accuracy, precision, recall, and F1 metrics. Accuracy metric enumerates the number of correct predicted changed and unchanged frames by TEEMs in early exit branches. Precision quantifies the number of frames which predicted as the changed frames and belong to the changed frame class. Recall quantifies the number of changed frame predictions made out of all changed frame examples.

 

Table \ref{tab:TEEMAccuracy} shows the measured accuracy, precision, recall and F1 for early exits of TEE-Faster-RCNN with Resnet50 and Resnet101 feature networks. TEE-Faster-RCNN is trained for the variation threshold of 0.4, equation \eqref{eq:SC}. Exit1, Exit2, Exit3, and Exit4 are located after the first, second, third, and fourth Bottleneck modules in the Faster-RCNN feature network, respectively. The Exit1 (with Resnet50 feature network) classifies frames into change and unchanged categories with 89\% accuracy. The classification accuracy increases up to 91\% and 93\% in Exit2 and Exit3. The observations show lower accuracy for Exit4 82\% compares to Exit3. We believe that lower accuracy of Exit4 is due to the low-resolution feature maps used by the  TEEM in early exit 4 to identify variations.

Exit4 uses $7\times7$ low-resolution but semantically strong feature, encoding very high-level concepts, to identify semantic variations across frames. Since the high-level features remain often unchanged across consecutive video frames, Exit4 achieves lower accuracy in classifying video frames with small semantic variation i.e., when the location of a small object changes slightly between two frames. However, Exit4 accurately detects unchanged frame pairs as well as frame pairs with significant variations. 

Compared to TEEMs, DFF \cite{shidifferential} detects the key-frames (refers to changed frames) with 76\% accuracy, yet with high computational cost. DFF concatenates the input frame pairs and sends them as a six-channel depth input through a differential network to get the difference feature map.
However, instead of processing input frame pairs together to identify variations, we proposed to process a video frame once and store and reuse the intermediate feature maps to build an attention map that encodes the semantic differences between frame pairs with small computational overhead.

\subsection{Class Activation Map of TEEMs}
\begin{figure*}[]
 \centering
 \footnotesize
\begin{tabular}{cccccc}
    \includegraphics[width=.14\linewidth]{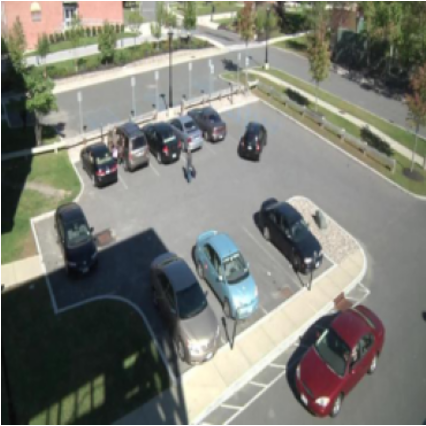} &  \includegraphics[width=.14\linewidth]{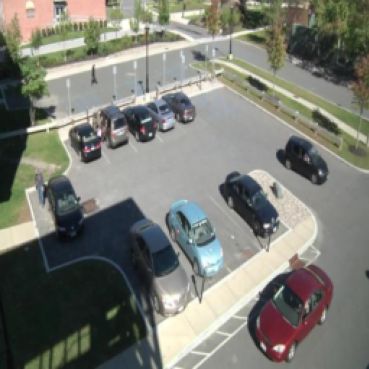} &       \includegraphics[width=.14\linewidth]{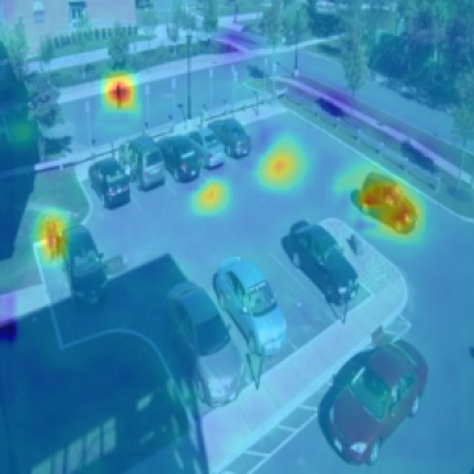} &       \includegraphics[width=.14\linewidth]{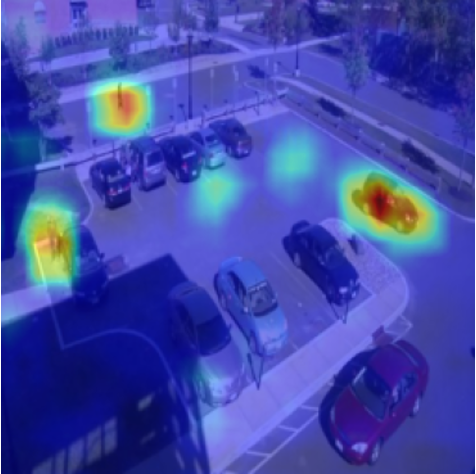} &       \includegraphics[width=.14\linewidth]{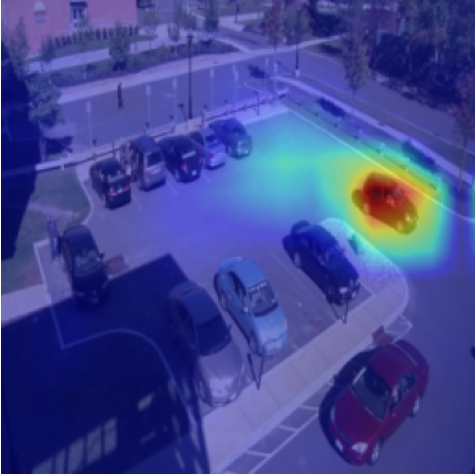} &     \includegraphics[width=.14\linewidth]{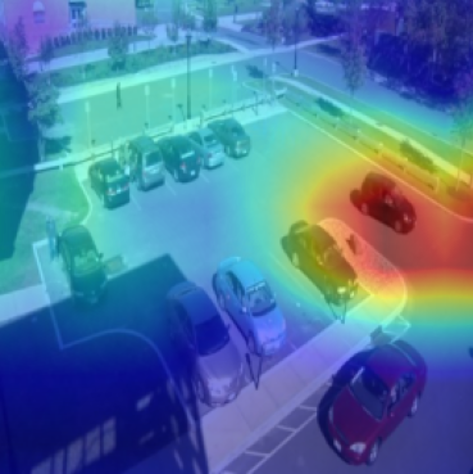}    \\
    \includegraphics[width=.14\linewidth]{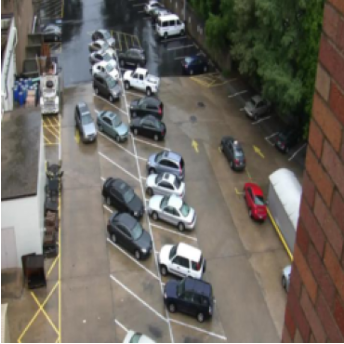} &  \includegraphics[width=.14\linewidth]{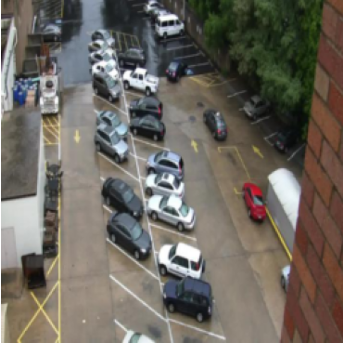} &       \includegraphics[width=.14\linewidth]{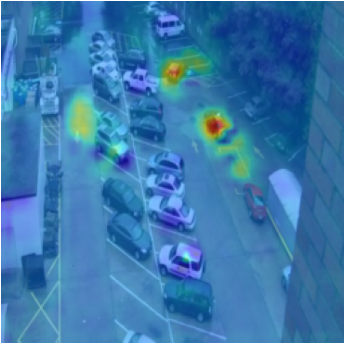} &       \includegraphics[width=.14\linewidth]{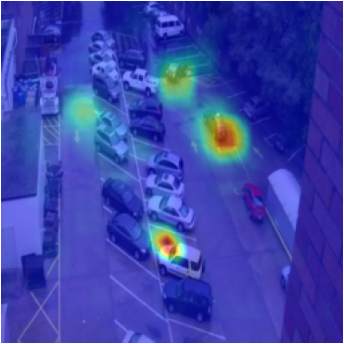} &       \includegraphics[width=.14\linewidth]{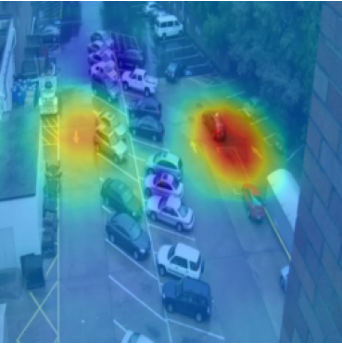} &     \includegraphics[width=.14\linewidth]{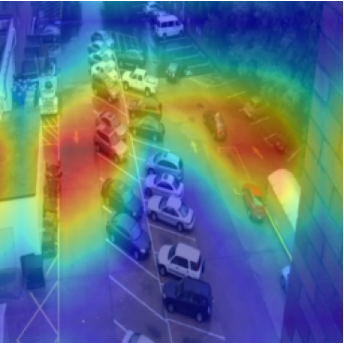}    \\
$fr_{i}$&    $fr_{i+j}$ &  $TEEM1$ &    $TEEM2$ &   $TEEM3$ &     $TEEM4$
\end{tabular}
\caption{Class activation maps of video frame pairs show that TEEMs effectively learn to focus only on the motions of the objects of interest between video frames to identify scenery change.}
\label{fig:CAM}
\end{figure*}

To visualize the performance of TEEMs, figure \ref{fig:CAM} shows class activation map of TEEMs. Each class activation map shows which parts of a video frame have contributed more to the final output of the TEEMs. Figure \ref{fig:CAM} shows that TEEMs effectively learn to focus on the moving fields of interest between frame pairs to identify variations. Futhermore, Figure \ref{fig:CAM} illustrates that TEEM1 and TEEM2 identify moving objects in higher resolution because the input feature maps into TEEM1 and TEEM2 have high resolution. The resolution of input feature maps into TEEM1 and TEEM2 are $56 \times 56 $ and $28 \times 28 $ resolution, respectively. However, the class activation map of TEEM3 and TEEM4 are coarse-grain because they use low-resolution feature maps of  $14 \times 14 $ and $7 \times 7 $, respectively.

\begin{table}[]
\footnotesize
\centering
\caption{Accuracy, precision, recall, and F1 scores measures for TEEMs in Faster-RCNN  with Resnet50 and Resnet101 feature networks.}
\begin{tabular}{@{}lcccccccc@{}}
\toprule
                  & \multicolumn{4}{c}{Resnet50}             & \multicolumn{4}{c}{Resnet101}            \\ \midrule
       &                 Acc &            Pr &          Re &          F1 &           Acc &          Pr &         Re &    F1 \\ \midrule
Exit1 &                  0.88 &          0.89 &        0.88&         0.89 &          0.89&          0.89 &      0.89  &  0.89        \\
Exit2 &                  0.93 &           0.93&        0.93&          0.93&          0.91&          0.92 &      0.92 &  0.92      \\
Exit3 &                  0.91 &           0.91&        0.92&          0.91&          0.94&          0.92 &      0.92 &   0.93       \\
Exit4 &                  0.86 &           0.82&        0.86&          0.84&          0.85&          0.83 &      0.87 &  0.85      \\ \bottomrule
\end{tabular}
    \label{tab:TEEMAccuracy}
\end{table}

\subsection{Computational Complexity}
Table \ref{tab:computation} compares computational cost (number of MAC operations and parameters) and run time (frames per second) of processing video frames by the original Faster-RCNN and  TEE-Faster-RCNN branches. The first row of table \ref{tab:computation} indicates the original Faster-RCNN network, including feature network, region proposal network, and region-based convolutional neural network.

Exit 1, Exit2, Exit3, and Exit4 refer to computational paths of TEE-Faster-RCNN from the input of Faster-RCNN to TEEM1, TEEM2, TEEM3, and TEEM4, respectively. The second column of table \ref{tab:computation} shows the required MAC operations and parameters for each computation path of TEE-Faster-RCNN with Resnet50 feature network. The third columns shows the speed of each early exit. The fourth column shows the computational complexity the TEEM added in early exits. The second part of table \ref{fig:overview} shows the same results for TEE-Faster-RCNN with resnet101 feature network.

The original Faster-RCNN with Resnet51 feature network requires up to 134G MAC operations and 41M parameters to process a video frame. The high computation requirement limits the video object detection speed to 18 fps, 14 fps for the Faster-RCNN with Resnet101 feature network. Using the same network architecture for processing all video frames regardless of the semantic variations between neighbouring frames leads to the inferior use of limited energy and computational resources. 

However, the TEE-Faster-RCNN uses computationally lightweight early exit branches to process unchanged video frames. Table \ref{fig:overview} shows that early exit branches have substantially less computations and memory requirements which speeds up processing video frames. Exit1 branch requires only 1.7G MAC operations and uses 0.5 M parameters. Significant reduction in computation complexity is because of avoided parts of the feature network, region proposal network, and region-based convolutional neural network. This reduction in computations speeds up processing unchanged video frames up to 628 fps. The required MAC operation for Exit2, Exit3 and Exit4 branches are 2.7G, 3.5G, and 5G, respectively. Exit2, Exit3 and Exit4 speed up processing unchanged frames to 390 fps, 263 fps, and 218 fps, respectively. The fourth column of table \ref{tab:computation}  shows the required MAC operation and number of parameters for TEEMs.

\begin{table*}[]
\caption{Computational complexity and speed of TEE-Faster-RCNN video object detection. }
\footnotesize
\centering
\begin{tabular}{@{}lcccccc@{}}
\toprule
            & \multicolumn{3}{c}{Resnet51}                     & \multicolumn{3}{c}{Resnet101}                      \\ \midrule
            & branch(Op, Par)      & Speed & TEEM(Opr,Par) & branch(Op, Par)        & Speed & TEEM(Opr,Par) \\ \midrule
FR & (134G , 41M)      & 18fps       & -           & (181G , 60) &    14fps          & -           \\
Exit1       & (1.7G , 0.525M) & 628fps      & (0.94G , 0.3M)   & (1.7G, 0.525M)   & 597fps         & (0.94G , 0.3M)   \\
Exit2       & (2.7G , 2.615M) & 390fps      & (0.93G , 1.17M)  & (3.7G, 2.910M)   & 400fps         & (0.93G , 1.17M)  \\
Exit3       & (3.5G , 9.733M) & 263fps       & (0.23G , 1.19M)  & (9.1G, 30.195M)  & 140fps         & (0.23G , 1.19M)  \\
Exit4       & (5G , 23.808)  & 218fps      & (0.25G , 5.129M)  & (9.9G, 50.289M)  & 126fps         & (0.2G , 5.129M)  \\ \bottomrule
\end{tabular}
\label{tab:computation}
\end{table*}

\subsection{Detection Accuracy}

Having tested the classification performance of TEEMs, we then evaluate the mean average precision (mAP @0.35:0.05:0.75) and mean intersection over union (mIOU)  of TEE-Faster-RCNN video object detection. Table \ref{table:acc} compares the accuracy of TEE-Faster-RCNN detection results with the per-frame original Faster-RCNN object detection. TEE-Faster-RCNN updates the detection results at the average step of 20. Therefore, for a fair comparison we performed Faster-RCNN with different fixed updating steps. The results reflect that the accuracy of TEE-Faster-RCNN cannot compete with the per-frame video object detection approach. Whilst, TEE-Faster-RCNN achieves the same accuracy of per-frame Faster-RCNN with the fixed updating steps of 7, its updating step of TEE-Faster-RCNN is 20 on average. Notably, TEE-Faster-RCNN does not aim to improve the accuracy of detection but introduces a simple yet effective approach to significantly reduce the computational complexity of video object detection for the video applications with less frequent moving objects e.g., the CDnet dataset \cite{CDnet}. For applications with frequent moving objects such as the ImageNet-VID dataset \cite{imagenet}, more complex methods like optical flow approaches are needed to achieve better accuracy.

\begin{table}[]
\centering
\caption{Comparing the detection accuracy of TEE-Faster-RCNN with per-frame Faster-RCNN}
\begin{tabular}{@{}llll@{}}
\toprule
            & Updating ratio & mAP  &mIoU \\ \midrule
Faster-RCNN & 1                 & 0.231    &0.8\\
Fixed-Step  &  7              &  0.211   &0.76\\
Fixed-Step  &  10              &  0.183   &0.75\\
Fixed-Step  &  20              &  0.16   &0.70\\
TEE-Faster-RCNN  &  20               & 0.209   &0.75 \\ \bottomrule
\end{tabular}
\label{table:acc}
\end{table}

\section{Conclusion}
We proposes a temporal early exit object detection pipeline to reduce the computational complexity of per-frame video object detection. The proposed approach takes advantage of infrequent variation between features of consecutive video frames to avoid redundant computation. Video frames with invariant features are identified in the early stages of the network with very low computation effort. For the unchanged video frames, detection results from previous frames are reused. A full computation effort is only required if a video frame is identified with semantic variations compared to previous frames. The proposed approach accelerates per-frame video object detection up to $34\times$ with less than 2.2 \% reduction in mAP.

\medskip
\bibliographystyle{plainnat}
\bibliography{neurips_2020.bib}

\end{document}